\documentclass[letterpaper]{article} 
\usepackage{aaai25}  
\usepackage{times}  
\usepackage{helvet}  
\usepackage{courier}  
\usepackage[hyphens]{url}  
\usepackage{graphicx} 
\usepackage{natbib}  
\usepackage{caption} 
\usepackage{algorithm}
\usepackage{algorithmic}

\usepackage{newfloat}
\usepackage{listings}
\DeclareCaptionStyle{ruled}{labelfont=normalfont,labelsep=colon,strut=off} 
\floatstyle{ruled}
\newfloat{listing}{tb}{lst}{}
\floatname{listing}{Listing}
\usepackage{amsmath}
\usepackage{booktabs}
\usepackage{multirow}

\usepackage{xcolor}
\newcommand{\answerYes}[1]{\textcolor{blue}{#1}} 
\newcommand{\answerNo}[1]{\textcolor{teal}{#1}} 
\newcommand{\answerNA}[1]{\textcolor{gray}{#1}}

\title{PRISM: Perceptual Recognition for Identifying Standout Moments in Human-Centric Keyframe Extraction}
\author {
    Mert Can Cakmak\textsuperscript{\rm 1},
    Nitin Agarwal\textsuperscript{\rm 1,2},
    Diwash Poudel\textsuperscript{\rm 1}
}
\affiliations {
    \textsuperscript{\rm 1}COSMOS Research Center, University of Arkansas, Little Rock, USA\\
    \textsuperscript{\rm 2}ICSI, University of California, Berkeley, USA\\
    \{mccakmak, nxagarwal, dpoudel\}@ualr.edu
}

\usepackage{bibentry}

\begin{document}

\maketitle

\begin{abstract}
Online videos play a central role in shaping political discourse and amplifying cyber social threats such as misinformation, propaganda, and radicalization. Detecting the most impactful or “standout” moments in video content is crucial for content moderation, summarization, and forensic analysis. In this paper, we introduce PRISM (Perceptual Recognition for Identifying Standout Moments), a lightweight and perceptually-aligned framework for keyframe extraction. PRISM operates in the CIELAB color space and uses perceptual color difference metrics to identify frames that align with human visual sensitivity. Unlike deep learning-based approaches, PRISM is interpretable, training-free, and computationally efficient, making it well suited for real-time and resource-constrained environments. We evaluate PRISM on four benchmark datasets: BBC, TVSum, SumMe, and ClipShots, and demonstrate that it achieves strong accuracy and fidelity while maintaining high compression ratios. These results highlight PRISM’s effectiveness in both structured and unstructured video content, and its potential as a scalable tool for analyzing and moderating harmful or politically sensitive media in online platforms.
\end{abstract}

%

\section{Introduction}

Online platforms like YouTube, TikTok, and X have become powerful engines for shaping public discourse, where video content spreads political narratives, social ideologies, and cyber threats at scale. Within this environment, identifying \textit{standout moments}, which are emotionally or ideologically charged video segments, is crucial for moderating harmful content and understanding its impact. Recent studies have examined how algorithmic bias and emotional content shape viewer exposure and engagement \cite{okeke2023examining, cakmak_analyzing_2024, cakmak_investigating_2024}.

These segments are often amplified through algorithmic curation and social sharing, and are frequently used to mislead or incite during sensitive events \cite{king2023diffusion, abdali2024multimodal, kaur2024deepfake}. As generative AI further obscures the line between real and synthetic media, detecting perceptually significant keyframes becomes vital for forensic analysis and misinformation detection. 

While NLP and network analysis are common in cyber threat research, visual signals in video remain underexplored, despite their strong emotional and narrative power \cite{amerini2025deepfake, seo2020visual, cakmak_examining_2025, bhattacharya2024analyzing, shajari2025developing}. Keyframe extraction helps surface these signals by condensing video into its most semantically relevant components, enabling tasks like moderation, summarization, and content indexing \cite{gurung_narratives_2025}.

\begin{figure}[!ht]
\centering
\includegraphics[width=0.9\columnwidth]{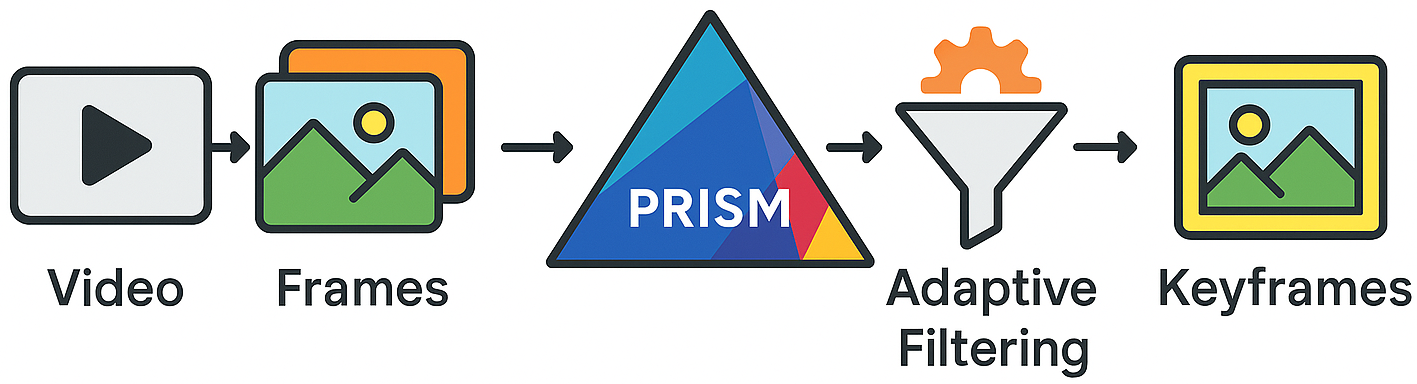} 
\caption{The PRISM framework for perceptually-guided keyframe extraction.}
\label{prism_flow}
\end{figure}

In this paper, we propose \textbf{PRISM} (Perceptual Recognition for Identifying Standout Moments), a lightweight, perceptually-driven framework for extracting visually meaningful frames. PRISM uses color-based perceptual metrics and adaptive filtering to identify standout content without training or heavy computation. As shown in Figure~\ref{prism_flow}, we evaluate PRISM across four diverse datasets and demonstrate its effectiveness in accuracy, fidelity, and compression, highlighting its potential for scalable and interpretable analysis of harmful media online.

\section{Literature Review}

The surge in video content across online platforms has driven demand for effective keyframe extraction methods \cite{alp_covid_2022}, essential for tasks such as summarization, indexing, and content retrieval. A variety of techniques have been proposed in recent years, each balancing different trade-offs between accuracy, complexity, and practical deployment.

Deep learning-based approaches have dominated the field, with frameworks like LMSKE~\cite{tan2024large} combining models such as TransNetV2 and CLIP for shot segmentation and semantic representation. Similarly, reinforcement learning methods~\cite{huang2022extracting} have been applied in medical imaging scenarios to extract diagnostically relevant frames. While these methods achieve high performance, they are often computationally intensive, require large annotated datasets, and are difficult to interpret or deploy in real-time systems.

Clustering-based methods~\cite{kaur2024effective}, such as those using fuzzy C-means with metaheuristics, offer a more lightweight alternative, but can suffer from sensitivity to initialization and a lack of perceptual alignment. Perceptual cues, including color-emotion links, have also been explored~\cite{cakmak_emotion_2024, yousefi_examining_2024}. Entropy and information-theoretic techniques~\cite{zhang2017key} attempt to select frames with high information content, though they may misfire in noisy or visually subtle transitions. Motion-based approaches~\cite{dong2022video} use temporal dynamics to locate changes but can miss static yet semantically important scenes. Object detection-based strategies~\cite{bharathi2023key} enhance semantic relevance but are constrained by category-specific detectors. Visual presentation factors, such as thumbnails and their design have also been shown to influence viewer engagement and algorithmic behavior~\cite{poudel2024beyond}. Bias in Shorts thumbnails and topic drift in recommendations have also been documented~\cite{cakmak_unveiling_2024, cakmak2025unpacking, cakmak2024bias}.

Despite this variety, common challenges persist across existing approaches, including poor interpretability, limited perceptual sensitivity, and high resource demands. Many techniques are ill-suited for real-time or large-scale applications where speed and simplicity are critical. Studies have also highlighted how multimedia plays a key role in shaping mobilization across sociopolitical contexts \cite{shaik_characterizing_2024}.

In contrast, we propose PRISM, a perceptually-guided, lightweight framework that selects standout frames using color differences in the CIELAB space. Unlike deep or heuristic systems, PRISM is training-free, interpretable, and efficient, making it practical for real-world use in constrained or high-throughput environments.

\section{Methodology}

In this research, we propose a perceptually-guided method for extracting standout keyframes from videos, using the CIELAB color space to better align with human visual perception.

\subsection{Justification of CIELAB Color Space}

Standard RGB spaces do not reflect human perception, as they are designed for hardware representation rather than perceptual uniformity. In contrast, CIELAB was specifically developed to align with human color sensitivity, where equal distances correspond to equal perceived differences. Using CIELAB thus enables more perceptually accurate analysis of color changes between frames.

\subsection{Frame Conversion Procedure}

The initial stage involves converting video frames from RGB color space (stored as BGR in OpenCV) to CIELAB color space. Mathematically, this transformation can be succinctly represented as

\begin{equation}
F_{LAB} = \text{cvtColor}(F_{RGB}, \text{COLOR\_BGR2LAB})
\end{equation}

This step ensures subsequent frame comparison computations reflect genuine perceptual differences, facilitating a robust and meaningful analysis of standout frames.

\subsection{Perceptual Color Difference Calculation}

We use the CIEDE2000 metric ($\Delta E_{00}$) \cite{pereira2019efficient} to measure color differences in a way that aligns with human perception. Unlike basic Euclidean distance, it accounts for variations in hue, chroma, and lightness. To simplify computation, we average the CIELAB values per frame and compute the perceptual difference accordingly.

\begin{equation}
\Delta E_{00}(F_i, F_{i+1}) = \text{CIEDE2000}\left(\overline{LAB}(F_i), \overline{LAB}(F_{i+1})\right)
\end{equation}

\subsection{Two-Stage Adaptive Thresholding for Keyframe Identification}

Our methodology utilizes an intricate two-stage adaptive thresholding mechanism specifically designed to distinguish significant perceptual changes from trivial variations.

\subsubsection{Initial Just Noticeable Difference (JND) Thresholding}

We first discard perceptual differences below the Just Noticeable Difference (JND), typically defined as a $\Delta E_{00}$ value of 1 \cite{chnspec_cie_lab}. Since CIEDE2000 is designed to reflect human color perception \cite{viewsonic_delta_e}, this threshold effectively filters out imperceptible variations from noise or lighting, allowing the analysis to focus on meaningful frame changes.

\begin{table}[!ht]
\small
\caption{Summary of $\Delta E$ Values and Their Interpretation}
\label{tab:deltaE_summary_short}
\centering
\begin{tabular}{@{}cl@{}}
\toprule
\textbf{$\Delta E$ Range} & \textbf{Meaning} \\ \midrule
0.0 -- 0.5   & Invisible to slight; acceptable \\
0.5 -- 2.0   & Noticeable on inspection; often acceptable \\
2.0 -- 4.0   & Clear difference; conditionally acceptable \\
4.0 -- 10.0  & Obvious difference; often unacceptable \\
$>$ 10.0     & Strong contrast; unacceptable \\ \bottomrule
\end{tabular}
\end{table}

\subsubsection{Adaptive Statistical Thresholding}

Following JND filtering, we apply adaptive statistical thresholding. Specifically, we calculate the mean ($\mu$) and standard deviation ($\sigma$) of the perceptually significant $\Delta E_{00}$ values within each video sequence as follows:

\begin{equation}
\mu = \frac{1}{N}\sum_{i=1}^{N} \Delta E_{00}(F_i, F_{i+1}), 
\end{equation}

\begin{equation}
 \sigma = \sqrt{\frac{1}{N}\sum_{i=1}^{N}\left(\Delta E_{00}(F_i, F_{i+1}) - \mu\right)^2}   
\end{equation}

Frames exhibiting perceptual differences exceeding the adaptive threshold ($\mu + \sigma$) are identified as standout frames or keyframes:

\begin{equation}
\Delta E_{00}(F_i, F_{i+1}) > \mu + \sigma
\end{equation}

This thresholding approach assumes a roughly Gaussian distribution of $\Delta E_{00}$ values. In such distributions, values exceeding $\mu + \sigma$ represent statistically significant deviations, commonly associated with meaningful outliers \cite{montgomery2007introduction}. This enables robust detection of perceptual changes while preserving relevant content.

Our method improves keyframe extraction by mimicking human perception and using adaptive thresholds to detect meaningful changes. It's a lightweight, efficient alternative to deep models, ideal for summarization and retrieval in resource-limited settings.
\section{Experiments and Evaluation}

This section evaluates the proposed PRISM framework using multiple benchmark datasets and a range of performance metrics to assess its accuracy, fidelity, and efficiency.

\subsection{Evaluation Datasets}

We evaluated the effectiveness of the proposed PRISM framework on a diverse set of publicly available video datasets that vary in length, content type, and annotation style. This allowed us to assess the model’s generalizability across different keyframe extraction scenarios, from user-generated clips to professionally edited documentaries.

The TVSum20 dataset \cite{tan2024large} contained 20 videos across 10 categories, each annotated with frame-level importance scores derived from crowdsourced feedback. SumMe \cite{gygli2014creating} included 25 user-generated videos, each paired with multiple human-generated summaries, enabling evaluation on subjective content. To test PRISM on long-form, high-resolution material, we used 10 selected episodes from the BBC Planet Earth series \cite{baraldi2016recognizing}, which featured rich visuals and complex transitions. We also included a subset of the ClipShots dataset \cite{tang2018fast}, composed of thousands of short videos with annotated shot boundaries, to evaluate PRISM's robustness in fast-paced and diverse video content.

These datasets provided a strong benchmark for evaluating PRISM’s scalability and efficiency, supported by prior work on large-scale transcript processing \cite{cakmak_adopting_2023, cakmak_high_2024}.

\subsection{Evaluation Metrics}

To fairly evaluate PRISM across videos of varying length and frame rate, we employed a dynamic frame-matching threshold that adapted to each video's properties. Fixed thresholds often disadvantage longer or high-fps videos, so our method scaled the tolerance window based on frame rate and total frame count.

As described in Algorithm~\ref{alg:fps-threshold}, a predicted keyframe is considered correct if it falls within a computed range of any ground-truth keyframe. The final score reflects the percentage of matched predictions.

\begin{algorithm}[ht]
\small
\caption{Frame Matching with FPS-Scaled Threshold}
\label{alg:fps-threshold}
\begin{algorithmic}[1]
\REQUIRE Row with \texttt{actual\_frame\_number}, \texttt{predicted\_frame\_number}, \texttt{frame\_count}, \texttt{fps}
\ENSURE Matching percentage

\STATE actual $\leftarrow$ row.actual\_frame\_number
\STATE predicted $\leftarrow$ row.predicted\_frame\_number
\STATE frame\_count $\leftarrow$ row.frame\_count
\STATE fps $\leftarrow$ row.fps

\IF{predicted is empty OR fps is 0 OR frame\_count is 0}
    \RETURN 0.0
\ENDIF

\STATE max\_time\_window $\leftarrow$ 10.0
\STATE alpha $\leftarrow$ 10
\STATE time\_scaling $\leftarrow$ max\_time\_window $\times \frac{fps}{fps + alpha}$
\STATE threshold $\leftarrow$ int(fps $\times$ time\_scaling)
\STATE min\_threshold $\leftarrow$ 30
\STATE max\_threshold $\leftarrow$ int(frame\_count $\times$ 0.03)
\STATE threshold $\leftarrow$ clamp(threshold, min\_threshold, max\_threshold)

\STATE match\_count $\leftarrow$ 0
\FORALL{pred in predicted}
    \IF{any $|$pred - act$|$ $\leq$ threshold for act in actual}
        \STATE match\_count $\leftarrow$ match\_count + 1
    \ENDIF
\ENDFOR

\RETURN round$\left(\frac{match\_count}{\text{len(predicted)}} \times 100, 2\right)$
\end{algorithmic}
\end{algorithm}

This adaptive strategy provided a balanced comparison across diverse datasets and video styles, from fast-paced clips to slow, documentary footage.

In addition to frame-matching accuracy (Algorithm~\ref{alg:fps-threshold}), we evaluated PRISM using two complementary metrics: \textit{fidelity} and \textit{compression ratio}, which capture representativeness and efficiency, respectively.

Fidelity measures how well the selected keyframes preserved the visual content of the original video. We computed it using cosine similarity between normalized color histograms of predicted and ground-truth keyframes, following the approach in \cite{tan2024large}. For each predicted frame, we found the most similar ground-truth frame, and defined the final score as

\begin{equation}
\text{Fidelity} = 1 - \max_{i} \left( \min_{j} \text{CosSim}(k_i, g_j) \right)
\end{equation}

Compression ratio is defined as

\begin{equation}
\text{Compression Ratio} = \frac{\text{Total frames}}{\text{Selected keyframes}}
\end{equation}

Higher fidelity indicates better content preservation, while a higher compression ratio reflects greater summarization efficiency. These metrics naturally trade off: fewer keyframes yield better compression but may reduce fidelity. PRISM aims to balance this trade-off by selecting a minimal yet perceptually meaningful set of frames.

\subsection{Evaluation Results}

Table~\ref{tab:results} presents PRISM’s performance across four benchmark video datasets, evaluated in terms of accuracy, fidelity, and compression ratio (CR). PRISM consistently achieves high compression ratios (above 98.8\%) while maintaining strong accuracy and perceptual fidelity.

\begin{table}[ht]
\small
\centering
\caption{Evaluation results of PRISM across four benchmark datasets. (CR: Compression Ratio)}
\label{tab:results}
\begin{tabular}{lccc}
\toprule
\textbf{Dataset} & \textbf{Accuracy (\%)} & \textbf{Fidelity (\%)} & \textbf{CR (\%)} \\
\midrule
BBC       & 99.50 & 75.70 & 99.50 \\
TVSum     & 78.40 & 74.85 & 99.20 \\
ClipShots & 66.24 & 68.31 & 99.42 \\
SumMe     & 96.16 & 64.35 & 98.80 \\
\bottomrule
\end{tabular}
\end{table}

The highest accuracy (99.50\%) and fidelity (75.70\%) were observed on the BBC dataset, which contains structured, long-form documentary footage, well-suited to PRISM’s perceptual approach. On TVSum, PRISM showed balanced performance (78.40\% accuracy, 74.85\% fidelity), reflecting its robustness across varied genres. In contrast, SumMe’s subjective, user-generated content led to a higher accuracy (96.16\%) but lower fidelity (64.35\%), revealing limitations when semantic importance is weakly correlated with visual cues.

ClipShots posed the greatest challenge due to fast editing and noise, yet PRISM still achieved reasonable fidelity (68.31\%) and the highest compression ratio (99.42\%). These results underscore PRISM’s effectiveness for both structured and unstructured content, validating its utility as a lightweight and generalizable keyframe extraction framework.

\begin{table}[ht]
\small
\centering
\caption{Average evaluation metrics for PRISM and other keyframe extraction models for the four datasets.}
\label{tab:avg_comparison}
\begin{tabular}{lccc}
\toprule
\textbf{Model} & \textbf{Accuracy (\%)} & \textbf{Fidelity (\%)} & \textbf{CR (\%)} \\
\midrule
\textbf{PRISM (Ours)}             & 85.58 & 70.30 & 99.23 \\
LiveLight        & 72.30 & 80.00 & 90.00 \\
DSVS & 66.00 & 75.00 & 95.00 \\
MKFE                     & 58.00 & —     & —     \\
VSLS  & 41.40 & —     & 98.60 \\
\bottomrule
\end{tabular}
\end{table}

To assess the overall effectiveness of PRISM, we compare its average performance across four benchmark datasets (TVSum, SumMe, ClipShots, and BBC Planet Earth) using three key metrics: accuracy, fidelity, and compression ratio (CR). As shown in Table~\ref{tab:avg_comparison}, PRISM achieves the highest average accuracy at 85.58\%, surpassing established methods such as LiveLight~\cite{zhao2014quasi} (72.30\%) and DSVS~\cite{cong2011towards} (66.00\%). It also offers an exceptional compression ratio of 99.23\%, significantly reducing video length while retaining critical content. Although PRISM's fidelity (70.30\%) is slightly lower than that of LiveLight (80.00\%), this reflects a deliberate trade-off favoring correctness and compactness over broader visual coverage. In scenarios where frame accuracy and compression are prioritized, such as indexing, summarization, or constrained environments, PRISM offers a highly effective solution.

Some values for models like MKFE~\cite{asim2018key} and VSLS~\cite{guo2025logic} are not available, as the original publications did not report fidelity or CR metrics. Nonetheless, the inclusion of these models provides historical and methodological context for comparing PRISM’s performance against both traditional and modern approaches.

\subsection{Time Complexity}

To assess the computational efficiency of PRISM, we compare its frame processing speed (FPS) with prior models across multiple datasets. Here, FPS refers to the number of video frames processed per second by the model, not the frame rate of the video itself. A higher FPS indicates faster analysis and better suitability for real-time or large-scale applications.

As shown in Table~\ref{tab:fps_comparison}, PRISM outperforms all baselines, achieving up to 454 FPS on ClipShots and over 130 FPS on other datasets, confirming its lightweight and efficient design.


\begin{table}[ht]
\scriptsize
\centering
\caption{FPS performance comparison of PRISM and prior models across benchmark datasets.}
\label{tab:fps_comparison}
\begin{tabular}{llc|llc}
\toprule
\textbf{Dataset} & \textbf{Model} & \textbf{FPS} & \textbf{Dataset} & \textbf{Model} & \textbf{FPS} \\
\midrule
\multirow{4}{*}{TVSum} 
    & \textbf{PRISM (Ours)} & 167 
    & \multirow{4}{*}{SumMe} 
    & \textbf{PRISM (Ours)} & 131 \\
& FIVT & 56 & & FIVT & 58 \\
& LiveLight & 25--30 & & LiveLight & 25--30 \\
& D-KTS & 15 & & D-KTS & 15 \\
\midrule
\multirow{5}{*}{ClipShots} 
    & \textbf{PRISM (Ours)} & 454 
    & \multirow{4}{*}{BBC} 
    & \textbf{PRISM (Ours)} & 185 \\
& DeepSBD & 382 & & TransNet V2 & 25--30 \\
& FastShot & 25 & & MFSBD & 31 \\
& TransNet V2 & 30 & & DASBD & 125 \\
& DASBD & 125 & & FFmpeg & 165 \\
\bottomrule
\end{tabular}
\end{table}

Compared models include FIVT~\cite{hsu2021video}, LiveLight~\cite{zhao2014quasi}, and D-KTS~\cite{ke2022towards}, as well as recent and competitive baselines like DeepSBD~\cite{tang2018fast}, TransNet V2~\cite{souvcek2019transnet}, FastShot~\cite{apostolidis2014fast}, DASBD~\cite{esteve2023shot}, and MFSBD~\cite{gushchin2021shot}. Many of these methods rely on deep learning or handcrafted features, which often introduce latency. In contrast, PRISM remains training-free and perceptually grounded, enabling real-time speeds without compromising performance.

These findings highlight PRISM’s practicality for deployment in systems requiring fast, interpretable, and scalable video analysis.

\section{Conclusion}

We introduced PRISM, a lightweight and perceptually-driven framework for keyframe extraction that operates without training or labeled data. By leveraging perceptual color differences in the CIELAB space, PRISM selects visually meaningful frames aligned with human perception.

Experiments across four diverse datasets show that PRISM achieves strong accuracy and fidelity while maintaining high compression, making it suitable for real-time and resource-constrained applications. PRISM also significantly outperforms prior methods in processing speed, confirming its practicality for large-scale and time-sensitive video analysis.

Future work will explore integrating semantic cues and temporal dynamics to further enhance performance in more complex or subjective video scenarios.

\section{Acknowledgments}

This research is funded in part by the U.S. National Science Foundation (OIA-1946391, OIA-1920920), U.S. Office of the Under Secretary of Defense for Research and Engineering (FA9550-22-1-0332), U.S. Army Research Office (W911NF-23-1-0011, W911NF-24-1-0078), U.S. Office of Naval Research (N00014-21-1-2121, N00014-21-1-2765, N00014-22-1-2318), U.S. Air Force Research Laboratory, U.S. Defense Advanced Research Projects Agency, the Australian Department of Defense Strategic Policy Grants Program, Arkansas Research Alliance, the Jerry L. Maulden/Entergy Endowment, and the Donaghey Foundation at the University of Arkansas at Little Rock. Any opinions, findings, and conclusions or recommendations expressed in this material are those of the authors and do not necessarily reflect the views of the funding organizations. The researchers gratefully acknowledge the support.

\bibliography{aaai25}

\begin{thebibliography}{47}
\providecommand{\natexlab}[1]{#1}

\bibitem[{Abdali, shaham, and Krishnamachari(2024)}]{abdali2024multimodal}
Abdali, S.; shaham, S.; and Krishnamachari, B. 2024.
\newblock Multi-modal Misinformation Detection: Approaches, Challenges and Opportunities.
\newblock arXiv:2203.13883.

\bibitem[{Alp et~al.(2022)Alp, Gergin, Eraslan, {\c{C}}akmak, and Alhajj}]{alp_covid_2022}
Alp, E.; Gergin, B.; Eraslan, Y.~A.; {\c{C}}akmak, M.~C.; and Alhajj, R. 2022.
\newblock \emph{Covid-19 and Vaccine Tweet Analysis}, 213--229.
\newblock Cham: Springer International Publishing.
\newblock ISBN 978-3-031-08242-9.

\bibitem[{Amerini et~al.(2025)Amerini, Barni, Battiato, Bestagini, Boato, Bruni, Caldelli, De~Natale, De~Nicola, Guarnera et~al.}]{amerini2025deepfake}
Amerini, I.; Barni, M.; Battiato, S.; Bestagini, P.; Boato, G.; Bruni, V.; Caldelli, R.; De~Natale, F.; De~Nicola, R.; Guarnera, L.; et~al. 2025.
\newblock Deepfake Media Forensics: Status and Future Challenges.
\newblock \emph{Journal of Imaging}, 11(3): 73.

\bibitem[{Apostolidis and Mezaris(2014)}]{apostolidis2014fast}
Apostolidis, E.; and Mezaris, V. 2014.
\newblock Fast shot segmentation combining global and local visual descriptors.
\newblock In \emph{2014 IEEE International Conference on Acoustics, Speech and Signal Processing (ICASSP)}, 6583--6587. IEEE.

\bibitem[{Asim et~al.(2018)Asim, Almaadeed, Al-M{\'a}adeed, Bouridane, and Beghdadi}]{asim2018key}
Asim, M.; Almaadeed, N.; Al-M{\'a}adeed, S.; Bouridane, A.; and Beghdadi, A. 2018.
\newblock A key frame based video summarization using color features.
\newblock In \emph{2018 Colour and visual computing symposium (CVCS)}, 1--6. IEEE.

\bibitem[{Baraldi, Grana, and Cucchiara(2016)}]{baraldi2016recognizing}
Baraldi, L.; Grana, C.; and Cucchiara, R. 2016.
\newblock Recognizing and presenting the storytelling video structure with deep multimodal networks.
\newblock \emph{IEEE Transactions on Multimedia}, 19(5): 955--968.

\bibitem[{Bharathi, Senthilarasi, and Hari(2023)}]{bharathi2023key}
Bharathi, S.; Senthilarasi, M.; and Hari, K. 2023.
\newblock Key frame extraction based on real-time person availability using YOLO.
\newblock \emph{Journal of Wireless Mobile Networks, Ubiquitous Computing, and Dependable Applications}, 14(2): 31--40.

\bibitem[{Bhattacharya, Agarwal, and Poudel(2024)}]{bhattacharya2024analyzing}
Bhattacharya, S.; Agarwal, N.; and Poudel, D. 2024.
\newblock Analyzing the impact of symbols in Taiwan’s election-related anti-disinformation campaign on TikTok.
\newblock \emph{Social Network Analysis and Mining}, 14(1): 1--20.

\bibitem[{Cakmak and Agarwal(2024)}]{cakmak_high_2024}
Cakmak, M.~C.; and Agarwal, N. 2024.
\newblock High-Speed Transcript Collection on Multimedia Platforms: Advancing Social Media Research through Parallel Processing.
\newblock In \emph{2024 IEEE International Parallel and Distributed Processing Symposium Workshops (IPDPSW)}, 857--860.

\bibitem[{Cakmak and Agarwal(2025)}]{cakmak2025unpacking}
Cakmak, M.~C.; and Agarwal, N. 2025.
\newblock Unpacking Algorithmic Bias in YouTube Shorts by Analyzing Thumbnails.
\newblock In \emph{Proceedings of the 58th Hawaii International Conference on System Sciences}.

\bibitem[{Cakmak et~al.(2024)Cakmak, Agarwal, Dagtas, and Poudel}]{cakmak_unveiling_2024}
Cakmak, M.~C.; Agarwal, N.; Dagtas, S.; and Poudel, D. 2024.
\newblock Unveiling Bias in YouTube Shorts: Analyzing Thumbnail Recommendations and Topic Dynamics.
\newblock In Thomson, R.; Hariharan, A.; Renshaw, S.; Al-khateeb, S.; Burger, A.; Park, P.; and Pyke, A., eds., \emph{Social, Cultural, and Behavioral Modeling}, 205--215. Cham: Springer Nature Switzerland.
\newblock ISBN 978-3-031-72241-7.

\bibitem[{Cakmak, Agarwal, and Oni(2024)}]{cakmak2024bias}
Cakmak, M.~C.; Agarwal, N.; and Oni, R. 2024.
\newblock The bias beneath: analyzing drift in YouTube’s algorithmic recommendations.
\newblock \emph{Social Network Analysis and Mining}, 14(1): 171.

\bibitem[{Cakmak et~al.(2025)Cakmak, Agarwal, Poudel, and Bhattacharya}]{cakmak_examining_2025}
Cakmak, M.~C.; Agarwal, N.; Poudel, D.; and Bhattacharya, S. 2025.
\newblock Examining the Impact of Symbolic Content on YouTube's Recommendation System.
\newblock In Cherifi, H.; Donduran, M.; Rocha, L.~M.; Cherifi, C.; and Varol, O., eds., \emph{Complex Networks {\&} Their Applications XIII}, 421--432. Cham: Springer Nature Switzerland.
\newblock ISBN 978-3-031-82431-9.

\bibitem[{Cakmak et~al.(2023{\natexlab{a}})Cakmak, Okeke, Onyepunuka, Spann, and Agarwal}]{cakmak_analyzing_2024}
Cakmak, M.~C.; Okeke, O.; Onyepunuka, U.; Spann, B.; and Agarwal, N. 2023{\natexlab{a}}.
\newblock Analyzing Bias in Recommender Systems: A Comprehensive Evaluation of YouTube's Recommendation Algorithm.
\newblock In \emph{Proceedings of the International Conference on Advances in Social Networks Analysis and Mining}, 753--760.

\bibitem[{Cakmak et~al.(2023{\natexlab{b}})Cakmak, Okeke, Onyepunuka, Spann, and Agarwal}]{cakmak_investigating_2024}
Cakmak, M.~C.; Okeke, O.; Onyepunuka, U.; Spann, B.; and Agarwal, N. 2023{\natexlab{b}}.
\newblock Investigating bias in YouTube recommendations: emotion, morality, and network dynamics in China-Uyghur content.
\newblock In \emph{International Conference on Complex Networks and Their Applications}, 351--362. Springer.

\bibitem[{Cakmak et~al.(2023{\natexlab{c}})Cakmak, Okeke, Spann, and Agarwal}]{cakmak_adopting_2023}
Cakmak, M.~C.; Okeke, O.; Spann, B.; and Agarwal, N. 2023{\natexlab{c}}.
\newblock Adopting Parallel Processing for Rapid Generation of Transcripts in Multimedia-rich Online Information Environment.
\newblock In \emph{2023 IEEE International Parallel and Distributed Processing Symposium Workshops (IPDPSW)}, 832--837.

\bibitem[{Cakmak, Shaik, and Agarwal(2024)}]{cakmak_emotion_2024}
Cakmak, M.~C.; Shaik, M.; and Agarwal, N. 2024.
\newblock Emotion assessment of youtube videos using color theory.
\newblock In \emph{Proceedings of the 2024 9th International Conference on Multimedia and Image Processing}, 6--14.

\bibitem[{{CHNSpec}(2024)}]{chnspec_cie_lab}
{CHNSpec}. 2024.
\newblock Why Use {CIE} Lab Color Metrics.
\newblock \url{https://www.chnspec.net/Why-Use-CIE-Lab-Color-Metrics.html}.
\newblock Accessed: 2025-03-05.

\bibitem[{Cong, Yuan, and Luo(2011)}]{cong2011towards}
Cong, Y.; Yuan, J.; and Luo, J. 2011.
\newblock Towards scalable summarization of consumer videos via sparse dictionary selection.
\newblock \emph{IEEE Transactions on Multimedia}, 14(1): 66--75.

\bibitem[{Dong et~al.(2022)Dong, Zhang, Zhang, and Zhang}]{dong2022video}
Dong, Y.; Zhang, Y.; Zhang, J.; and Zhang, X. 2022.
\newblock Video key frame extraction based on scale and direction analysis.
\newblock \emph{The Journal of Engineering}, 2022(9): 910--918.

\bibitem[{Esteve~Brotons et~al.(2023)Esteve~Brotons, Lucendo, Javier, and Garcia-Rodriguez}]{esteve2023shot}
Esteve~Brotons, M.~J.; Lucendo, F.~J.; Javier, R.-J.; and Garcia-Rodriguez, J. 2023.
\newblock Shot Boundary Detection with 3D Depthwise Convolutions and Visual Attention.
\newblock \emph{Sensors}, 23(16): 7022.

\bibitem[{{FORCE11}(2020)}]{fair}
{FORCE11}. 2020.
\newblock The FAIR Data principles.
\newblock \url{https://force11.org/info/the-fair-data-principles/}.

\bibitem[{Gebru et~al.(2021)Gebru, Morgenstern, Vecchione, Vaughan, Wallach, Iii, and Crawford}]{gebru2021datasheets}
Gebru, T.; Morgenstern, J.; Vecchione, B.; Vaughan, J.~W.; Wallach, H.; Iii, H.~D.; and Crawford, K. 2021.
\newblock Datasheets for datasets.
\newblock \emph{Communications of the ACM}, 64(12): 86--92.

\bibitem[{Guo et~al.(2025)Guo, Chen, Wang, He, Xu, Ye, Sun, and Xiong}]{guo2025logic}
Guo, W.; Chen, Z.; Wang, S.; He, J.; Xu, Y.; Ye, J.; Sun, Y.; and Xiong, H. 2025.
\newblock Logic-in-Frames: Dynamic Keyframe Search via Visual Semantic-Logical Verification for Long Video Understanding.
\newblock \emph{arXiv preprint arXiv:2503.13139}.

\bibitem[{Gurung, Agarwal, and Al-Taweel(2025)}]{gurung_narratives_2025}
Gurung, M.~I.; Agarwal, N.; and Al-Taweel, A. 2025.
\newblock Are Narratives Contagious? Modeling Narrative Diffusion Using Epidemiological Theories.
\newblock In Aiello, L.~M.; Chakraborty, T.; and Gaito, S., eds., \emph{Social Networks Analysis and Mining}, 303--318. Cham: Springer Nature Switzerland.
\newblock ISBN 978-3-031-78554-2.

\bibitem[{Gushchin, Antsiferova, and Vatolin(2021)}]{gushchin2021shot}
Gushchin, A.; Antsiferova, A.; and Vatolin, D. 2021.
\newblock Shot boundary detection method based on a new extensive dataset and mixed features.
\newblock \emph{arXiv preprint arXiv:2109.01057}.

\bibitem[{Gygli et~al.(2014)Gygli, Grabner, Riemenschneider, and Van~Gool}]{gygli2014creating}
Gygli, M.; Grabner, H.; Riemenschneider, H.; and Van~Gool, L. 2014.
\newblock Creating summaries from user videos.
\newblock In \emph{Computer Vision--ECCV 2014: 13th European Conference, Zurich, Switzerland, September 6-12, 2014, Proceedings, Part VII 13}, 505--520. Springer.

\bibitem[{Hsu, Liao, and Huang(2021)}]{hsu2021video}
Hsu, T.-C.; Liao, Y.-S.; and Huang, C.-R. 2021.
\newblock Video summarization with frame index vision transformer.
\newblock In \emph{2021 17th International Conference on Machine Vision and Applications (MVA)}, 1--5. IEEE.

\bibitem[{Huang et~al.(2022)Huang, Ying, Lin, Zheng, Tan, Tang, Zhang, Luo, Yi, Liu et~al.}]{huang2022extracting}
Huang, R.; Ying, Q.; Lin, Z.; Zheng, Z.; Tan, L.; Tang, G.; Zhang, Q.; Luo, M.; Yi, X.; Liu, P.; et~al. 2022.
\newblock Extracting keyframes of breast ultrasound video using deep reinforcement learning.
\newblock \emph{Medical image analysis}, 80: 102490.

\bibitem[{Kaur et~al.(2024)Kaur, Noori~Hoshyar, Saikrishna, Firmin, and Xia}]{kaur2024deepfake}
Kaur, A.; Noori~Hoshyar, A.; Saikrishna, V.; Firmin, S.; and Xia, F. 2024.
\newblock Deepfake video detection: challenges and opportunities.
\newblock \emph{Artificial Intelligence Review}, 57(6): 159.

\bibitem[{Kaur, Kaur, and Lal(2024)}]{kaur2024effective}
Kaur, S.; Kaur, L.; and Lal, M. 2024.
\newblock An effective Key Frame Extraction technique based on Feature Fusion and Fuzzy-C means clustering with Artificial Hummingbird.
\newblock \emph{Scientific Reports}, 14(1): 26651.

\bibitem[{Ke et~al.(2022)Ke, Chang, Wu, Xu, and Zhong}]{ke2022towards}
Ke, X.; Chang, B.; Wu, H.; Xu, F.; and Zhong, S. 2022.
\newblock Towards practical and efficient long video summary.
\newblock In \emph{ICASSP 2022-2022 IEEE International Conference on Acoustics, Speech and Signal Processing (ICASSP)}, 1770--1774. IEEE.

\bibitem[{King and Wang(2023)}]{king2023diffusion}
King, K.~K.; and Wang, B. 2023.
\newblock Diffusion of real versus misinformation during a crisis event: A big data-driven approach.
\newblock \emph{International Journal of Information Management}, 71: 102390.

\bibitem[{Montgomery(2007)}]{montgomery2007introduction}
Montgomery, D.~C. 2007.
\newblock \emph{Introduction to Statistical Quality Control}.
\newblock John Wiley \& Sons.

\bibitem[{Okeke et~al.(2023)Okeke, Cakmak, Spann, and Agarwal}]{okeke2023examining}
Okeke, O.; Cakmak, M.~C.; Spann, B.; and Agarwal, N. 2023.
\newblock Examining content and emotion bias in youtube’s recommendation algorithm.
\newblock In \emph{the Ninth International Conference on Human and Social Analytics, Barcelona, Spain}.

\bibitem[{Pereira et~al.(2019)Pereira, Carvalho, Coelho, and C{\^o}rte-Real}]{pereira2019efficient}
Pereira, A.; Carvalho, P.; Coelho, G.; and C{\^o}rte-Real, L. 2019.
\newblock Efficient CIEDE2000-based color similarity decision for computer vision.
\newblock \emph{IEEE Transactions on Circuits and Systems for Video Technology}, 30(7): 2141--2154.

\bibitem[{Poudel, Cakmak, and Agarwal(2024)}]{poudel2024beyond}
Poudel, D.; Cakmak, M.~C.; and Agarwal, N. 2024.
\newblock Beyond the click: How youtube thumbnails shape user interaction and algorithmic recommendations.
\newblock In \emph{The 16th International Conference on Advances in Social Networks Analysis and Mining (ASONAM), accepted for presentation}.

\bibitem[{Seo(2020)}]{seo2020visual}
Seo, H. 2020.
\newblock Visual propaganda and social media.
\newblock \emph{Handbook of Propaganda}, 126--137.

\bibitem[{Shaik et~al.(2024)Shaik, Cakmak, Spann, and Agarwal}]{shaik_characterizing_2024}
Shaik, M.; Cakmak, M.~C.; Spann, B.; and Agarwal, N. 2024.
\newblock Characterizing Multimedia Adoption and its Role on Mobilization in Social Movements.
\newblock In \emph{Proceedings of the 57th Hawaii International Conference on System Sciences}.

\bibitem[{Shajari and Agarwal(2025)}]{shajari2025developing}
Shajari, S.; and Agarwal, N. 2025.
\newblock Developing a network-centric approach for anomalous behavior detection on youtube.
\newblock \emph{Social Network Analysis and Mining}, 15(1): 3.

\bibitem[{Sou{\v{c}}ek, Moravec, and Loko{\v{c}}(2019)}]{souvcek2019transnet}
Sou{\v{c}}ek, T.; Moravec, J.; and Loko{\v{c}}, J. 2019.
\newblock Transnet: A deep network for fast detection of common shot transitions.
\newblock \emph{arXiv preprint arXiv:1906.03363}.

\bibitem[{Tan et~al.(2024)Tan, Zhou, Xia, Liu, and Chen}]{tan2024large}
Tan, K.; Zhou, Y.; Xia, Q.; Liu, R.; and Chen, Y. 2024.
\newblock Large model based sequential keyframe extraction for video summarization.
\newblock In \emph{Proceedings of the International Conference on Computing, Machine Learning and Data Science}, 1--5.

\bibitem[{Tang et~al.(2018)Tang, Feng, Kuang, Chen, and Zhang}]{tang2018fast}
Tang, S.; Feng, L.; Kuang, Z.; Chen, Y.; and Zhang, W. 2018.
\newblock Fast video shot transition localization with deep structured models.
\newblock In \emph{Asian Conference on Computer Vision}, 577--592. Springer.

\bibitem[{{ViewSonic}(2021)}]{viewsonic_delta_e}
{ViewSonic}. 2021.
\newblock What is Delta E? And Why Is It Important for Color Accuracy?
\newblock \url{https://www.viewsonic.com/library/creative-work/what-is-delta-e-and-why-is-it-important-for-color-accuracy/}.
\newblock Accessed: 2025-03-05.

\bibitem[{Yousefi, Cakmak, and Agarwal(2024)}]{yousefi_examining_2024}
Yousefi, N.; Cakmak, M.~C.; and Agarwal, N. 2024.
\newblock Examining Multimodel Emotion Assessment and Resonance with Audience on YouTube.
\newblock In \emph{Proceedings of the 2024 9th International Conference on Multimedia and Image Processing}, ICMIP '24, 85–93. New York, NY, USA: Association for Computing Machinery.
\newblock ISBN 9798400716164.

\bibitem[{Zhang, Tian, and Li(2017)}]{zhang2017key}
Zhang, M.; Tian, L.; and Li, C. 2017.
\newblock Key frame extraction based on entropy difference and perceptual hash.
\newblock In \emph{2017 IEEE International Symposium on Multimedia (ISM)}, 557--560. IEEE.

\bibitem[{Zhao and Xing(2014)}]{zhao2014quasi}
Zhao, B.; and Xing, E.~P. 2014.
\newblock Quasi real-time summarization for consumer videos.
\newblock In \emph{Proceedings of the IEEE conference on computer vision and pattern recognition}, 2513--2520.

\end{thebibliography}

\subsection{Paper Checklist}

\begin{enumerate}

\item For most authors...
\begin{enumerate}
    \item Would answering this research question advance science without violating social contracts, such as violating privacy norms, perpetuating unfair profiling, exacerbating the socio-economic divide, or implying disrespect to societies or cultures?
    \answerYes{Yes, we use public video data without involving private or sensitive information.}
    
    \item Do your main claims in the abstract and introduction accurately reflect the paper's contributions and scope?
\answerYes{Yes, they clearly reflect PRISM’s contributions and scope.}
    
    \item Do you clarify how the proposed methodological approach is appropriate for the claims made? 
\answerYes{Yes, Sections 3 and 4 justify the method and evaluation.}
    
    \item Do you clarify what are possible artifacts in the data used, given population-specific distributions?
\answerNo{No, the datasets are public benchmarks without demographic or sensitive data.}
    \item Did you describe the limitations of your work?
\answerYes{Yes, discussed in Section 6.}
    
    \item Did you discuss any potential negative societal impacts of your work?
\answerYes{Yes, discussed in the Introduction and Conclusion.}
    
    \item Did you discuss any potential misuse of your work?
\answerYes{Yes, we address potential misuse and responsible deployment in the Conclusion.}
    
    \item Did you describe steps taken to prevent or mitigate potential negative outcomes of the research, such as data and model documentation, data anonymization, responsible release, access control, and the reproducibility of findings?
\answerYes{We use public data, no training, and detail our reproducible setup.}
    \item Have you read the ethics review guidelines and ensured that your paper conforms to them?
\answerYes{Yes, the paper follows ethical guidelines and uses no private data.}
\end{enumerate}

\item Additionally, if your study involves hypotheses testing...
\begin{enumerate}
  \item Did you clearly state the assumptions underlying all theoretical results?
    \answerNA{NA}
  \item Have you provided justifications for all theoretical results?
    \answerNA{NA}
  \item Did you discuss competing hypotheses or theories that might challenge or complement your theoretical results?
    \answerNA{NA}
  \item Have you considered alternative mechanisms or explanations that might account for the same outcomes observed in your study?
    \answerNA{NA}
  \item Did you address potential biases or limitations in your theoretical framework?
    \answerNA{NA}
  \item Have you related your theoretical results to the existing literature in social science?
    \answerNA{NA}
  \item Did you discuss the implications of your theoretical results for policy, practice, or further research in the social science domain?
    \answerNA{NA}
\end{enumerate}

\item Additionally, if you are including theoretical proofs...
\begin{enumerate}
  \item Did you state the full set of assumptions of all theoretical results?
    \answerNA{NA}
  \item Did you include complete proofs of all theoretical results?
    \answerNA{NA}
\end{enumerate}

\item Additionally, if you ran machine learning experiments...
\begin{enumerate}
  \item Did you include the code, data, and instructions needed to reproduce the main experimental results (either in the supplemental material or as a URL)?
    \answerNA{NA}
    
  \item Did you specify all the training details (e.g., data splits, hyperparameters, how they were chosen)?
    \answerNA{NA}
    
  \item Did you report error bars (e.g., with respect to the random seed after running experiments multiple times)?
    \answerNA{NA}
    
  \item Did you include the total amount of compute and the type of resources used (e.g., type of GPUs, internal cluster, or cloud provider)?
    \answerNA{NA}
    
  \item Do you justify how the proposed evaluation is sufficient and appropriate to the claims made? 
    \answerNA{NA}
    
  \item Do you discuss what is ``the cost`` of misclassification and fault (in)tolerance?
    \answerNA{NA}
\end{enumerate}

\item Additionally, if you are using existing assets (e.g., code, data, models) or curating/releasing new assets, \textbf{without compromising anonymity}...
\begin{enumerate}
  \item If your work uses existing assets, did you cite the creators?
\answerYes{Yes, all datasets are cited in the Evaluation section.}
    
  \item Did you mention the license of the assets?
    \answerNo{No, but we only use publicly available academic datasets.}
    
  \item Did you include any new assets in the supplemental material or as a URL?
\answerNo{No, we included no new assets or links.}

  \item Did you discuss whether and how consent was obtained from people whose data you're using/curating?
    \answerNA{NA – no personal or user data is used.}
    
  \item Did you discuss whether the data you are using/curating contains personally identifiable information or offensive content?
\answerYes{Yes, the public datasets contain no PII or offensive content.}
    
  \item If you are curating or releasing new datasets, did you discuss how you intend to make your datasets FAIR (see \citet{fair})?
    \answerNA{NA}
    
  \item If you are curating or releasing new datasets, did you create a Datasheet for the Dataset (see \citet{gebru2021datasheets})? 
    \answerNA{NA}
\end{enumerate}

\item Additionally, if you used crowdsourcing or conducted research with human subjects, \textbf{without compromising anonymity}...
\begin{enumerate}
  \item Did you include the full text of instructions given to participants and screenshots?
    \answerNA{NA}
    
  \item Did you describe any potential participant risks, with mentions of Institutional Review Board (IRB) approvals?
    \answerNA{NA}
    
  \item Did you include the estimated hourly wage paid to participants and the total amount spent on participant compensation?
    \answerNA{NA}
    
  \item Did you discuss how data is stored, shared, and deidentified?
    \answerNA{NA}
\end{enumerate}

\end{enumerate}

\end{document}